\documentclass[letterpaper]{article} % DO NOT CHANGE THIS

\usepackage{aaai2026}              % Camera-ready version

\usepackage{times}  % DO NOT CHANGE THIS
\usepackage{helvet}  % DO NOT CHANGE THIS
\usepackage{courier}  % DO NOT CHANGE THIS
\usepackage[hyphens]{url}  % DO NOT CHANGE THIS
\usepackage{graphicx} % DO NOT CHANGE THIS
\usepackage{natbib}  % DO NOT CHANGE THIS AND DO NOT ADD ANY OPTIONS TO IT
\usepackage{caption} % DO NOT CHANGE THIS AND DO NOT ADD ANY OPTIONS TO IT
\usepackage{amsmath}
\usepackage{amsfonts}
\usepackage{amssymb}
\usepackage{booktabs}      % 支持 \toprule, \midrule, \bottomrule
\usepackage{multirow}      % 支持 \multirow
\usepackage{graphicx}      % 支持 \resizebox
\usepackage{array}         % 支持列宽设置
\usepackage{caption}       % 控制标题位置
\usepackage{amssymb}
\usepackage{pifont}

\usepackage{color}
\usepackage[table,xcdraw,dvipsnames]{xcolor}
\definecolor{myPink}{rgb}{0.9294, 0.0078, 0.5490}
\definecolor{Gray}{gray}{0.92}
\definecolor{mygray}{rgb}{.95,.95,.95}

\definecolor{lightgray}{gray}{0.9}
\definecolor{lightblue}{rgb}{0.68, 0.85, 0.9}

\definecolor{lightpurple}{rgb}{0.85, 0.75, 0.95}
\definecolor{lightorange}{rgb}{1.0, 0.9, 0.8}

\definecolor{my_color}{HTML}{E8F3F1}

\usepackage{algorithm}
\usepackage{algorithmic}

\usepackage{newfloat}
\usepackage{listings}
\DeclareCaptionStyle{ruled}{labelfont=normalfont,labelsep=colon,strut=off} % DO NOT CHANGE THIS
\floatstyle{ruled}
\newfloat{listing}{tb}{lst}{}
\floatname{listing}{Listing}

\title{ClearAIR: A Human-Visual-Perception-Inspired All-in-One Image Restoration}

\author{
    Xu Zhang\textsuperscript{\rm 1}, Huan Zhang\textsuperscript{\rm 2},
    Guoli Wang\textsuperscript{\rm 3}, Qian Zhang\textsuperscript{\rm 3},
    Lefei Zhang\textsuperscript{\rm 1}\thanks{Corresponding author.}
}

\affiliations{
    \textsuperscript{\rm 1}National Engineering Research Center for Multimedia Software, School of Computer Science, Wuhan University \\
    \textsuperscript{\rm 2}School of Information Engineering, Guangdong University of Technology\\
    \textsuperscript{\rm 3}Horizon Robotics
    \\
    \{zhangx0802, zhanglefei\}@whu.edu.cn, huanzhang2021@gdut.edu.cn, \{guoli.wang, qian01.zhang\}@horizon.auto
}

\usepackage{bibentry}
\begin{document}

\maketitle

\begin{abstract}

All-in-One Image Restoration (AiOIR) has advanced significantly, offering promising solutions for complex real-world degradations. However, most existing approaches rely heavily on degradation-specific representations, often resulting in oversmoothing and artifacts. To address this, we propose ClearAIR, a novel AiOIR framework inspired by Human Visual Perception (HVP) and designed with a hierarchical, coarse-to-fine restoration strategy.
First, leveraging the global priority of early HVP, we employ a Multimodal Large Language Model (MLLM)-based Image Quality Assessment (IQA) model for overall evaluation. Unlike conventional IQA, our method integrates cross-modal understanding to more accurately characterize complex, composite degradations.
Building upon this overall assessment, we then introduce a region awareness and task recognition pipeline. A semantic cross-attention, leveraging semantic guidance unit, first produces coarse semantic prompts. Guided by this regional context, a degradation-aware module implicitly captures region-specific degradation characteristics, enabling more precise local restoration.
Finally, to recover fine details, we propose an internal clue reuse mechanism. It operates in a self-supervised manner to mine and leverage the intrinsic information of the image itself, substantially enhancing detail restoration.
Experimental results show that ClearAIR achieves superior performance across diverse synthetic and real-world datasets.
\end{abstract}

% % Links section - only shown in camera-ready version
% \ifdefined\aaaianonymous
% % Uncomment the following to link to your code, datasets, an extended version or similar.
% % You must keep this block between (not within) the abstract and the main body of the paper.
% % NOTE: For anonymous submissions, do not include links that could reveal your identity
% % \begin{links}
% %     \link{Code}{https://aaai.org/example/code}
% %     \link{Datasets}{https://aaai.org/example/datasets}
% %     \link{Extended version}{https://aaai.org/example/extended-version}
% % \end{links}
% \else
% % Uncomment the following to link to your code, datasets, an extended version or similar.
% % You must keep this block between (not within) the abstract and the main body of the paper.
% \begin{links}
%     \link{Code}{https://github.com/House-yuyu/ClearAIR}
%     % \link{Datasets}{https://aaai.org/example/datasets}
%     % \link{Extended version}{https://aaai.org/example/extended-version}
% \end{links}
% \fi

\section{Introduction}
\label{Introduction}

Image restoration aims to recover a clean image from its degraded version and has made significant progress with deep learning. Early approaches employed task-specific networks for individual degradation types, such as denoising \cite{BSD68}, dehazing \cite{SYK_haze, SOTS}, deraining \cite{MSPFN}, deblurring \cite{GoPro, SAM_blur}, and low-light enhancement \cite{LOL}, achieving strong performance within their intended domains. However, these methods lack generalization across tasks. Although general-purpose restoration models \cite{NAFNet, Restormer} have been developed to handle multiple degradations, they often still require separate models for each degradation type, resulting in complex inference and increased computational costs.

In recent years, All-in-One Image Restoration (AiOIR) methods \cite{wu_24AAAI, Foundir} have emerged as promising solutions. These frameworks can simultaneously handle diverse degradation types through various mechanisms. Early efforts, like AirNet \cite{AirNet}, focused on creating specific degradation encoders to capture distinctive feature representations. Subsequent innovations, such as those found in ProRes \cite{ProRes} and PromptIR \cite{PromptIR}, enhanced performance by incorporating visual prompts. More recent research \cite{MPerceiver, PerceiveIR} has harnessed the powerful feature extraction capabilities of large-scale visual models to improve texture reconstruction and ensure structural integrity. 
However, these AiOIR methods overlook a critical issue: spatially non-uniform degradations can significantly alter the local statistical properties of an image. Most existing AiOIR approaches apply a uniform processing strategy across the entire image, failing to account for variations in degradation distribution and severity across different regions.

% ***********************************************************
\begin{figure*}[!htb]
\centering
\includegraphics[width=0.92\textwidth]{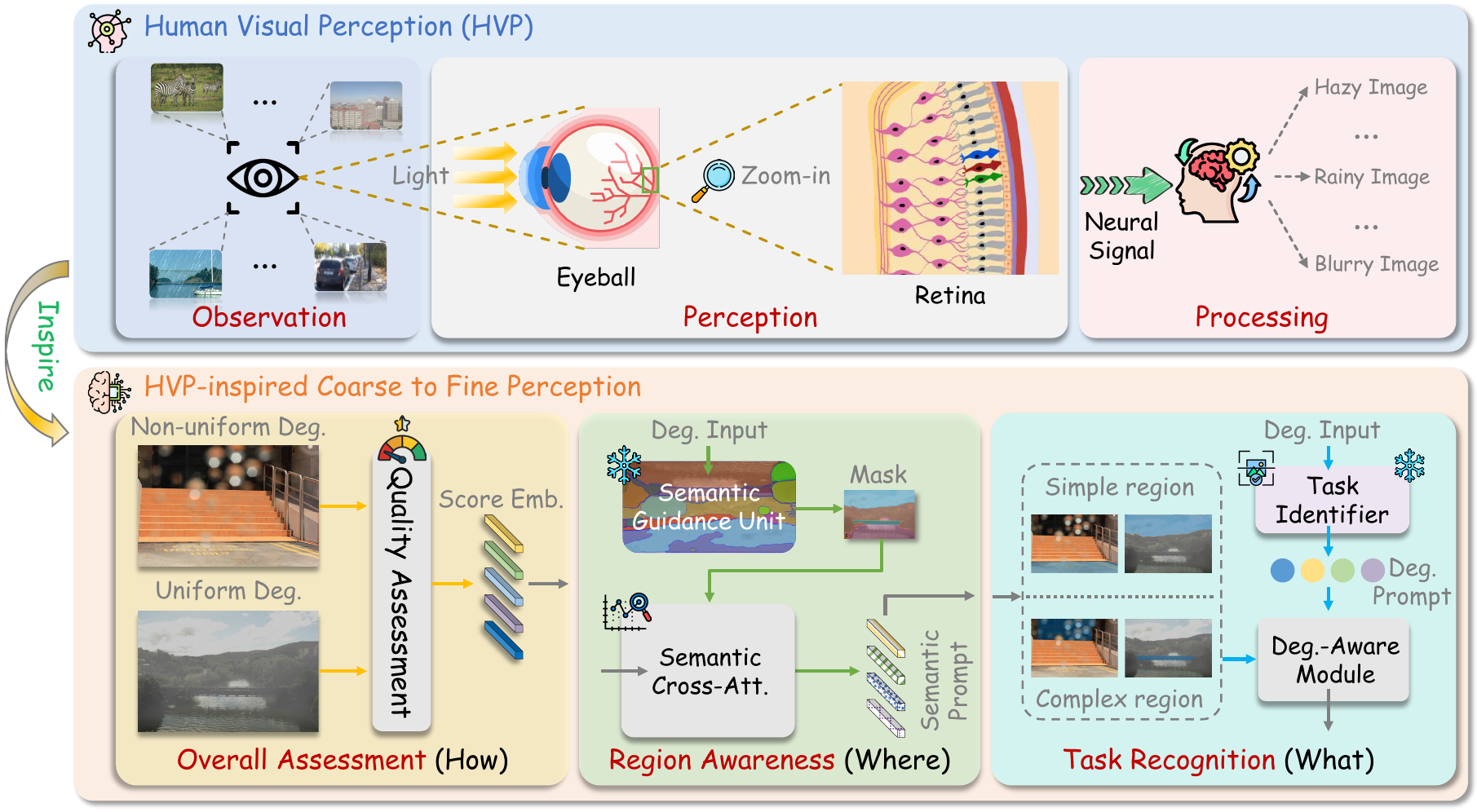}
% \vspace{-2mm}
\caption{
A coarse-to-fine image processing pipeline inspired by human visual perception.
}
\label{fig:motivation}
\end{figure*}

To address this limitation, as shown in Fig.~\ref{fig:motivation}, we design a progressive restoration pipeline inspired by Human Visual Perception (HVP), which refines image quality hierarchically from global structure to fine local details.
First, as in early HVP stages, which emphasize global structure, we integrate an MLLM-based Image Quality Assessment (IQA) model to evaluate the image’s overall quality.
Second, to better account for spatially varying degradation patterns, we incorporate a Semantic Guidance Unit (SGU) to support region-level segmentation and provide coarse guidance for identifying areas likely affected by degradation.
Third, guided by the spatial cues from the SGU, we apply a task identifier to estimate the predominant degradation type in local neighborhoods. This allows ClearAIR to adaptively select region-appropriate restoration strategies, avoiding a uniform one-size-fits-all treatment across the image.
Finally, to enhance the recovery of fine-grained local details, we propose an Internal Clue Reuse Mechanism (ICRM) that leverages internal image statistics to refine local structures.

Our main contributions can be summarized as follows:

\begin{itemize}
    \item 
    We present ClearAIR, a novel AiOIR framework inspired by HVP. By adopting a coarse-to-fine hierarchical restoration process, it gradually improves both structural integrity and perceptual quality.
    \item 
    We propose an HVP-inspired pipeline integrating global quality and local semantic cues. An MLLM-based IQA evaluates quality, while the SGU and task identifier guide regional analysis and degradation estimation.
    \item 
    We introduce ICRM that leverages self-supervised learning to exploit the intrinsic structure of the image, thereby enhancing the model’s ability to recover fine local details.
\end{itemize}

% ***********************************************************
\begin{figure*}[!htb]
\centering
\includegraphics[width=0.98\textwidth]{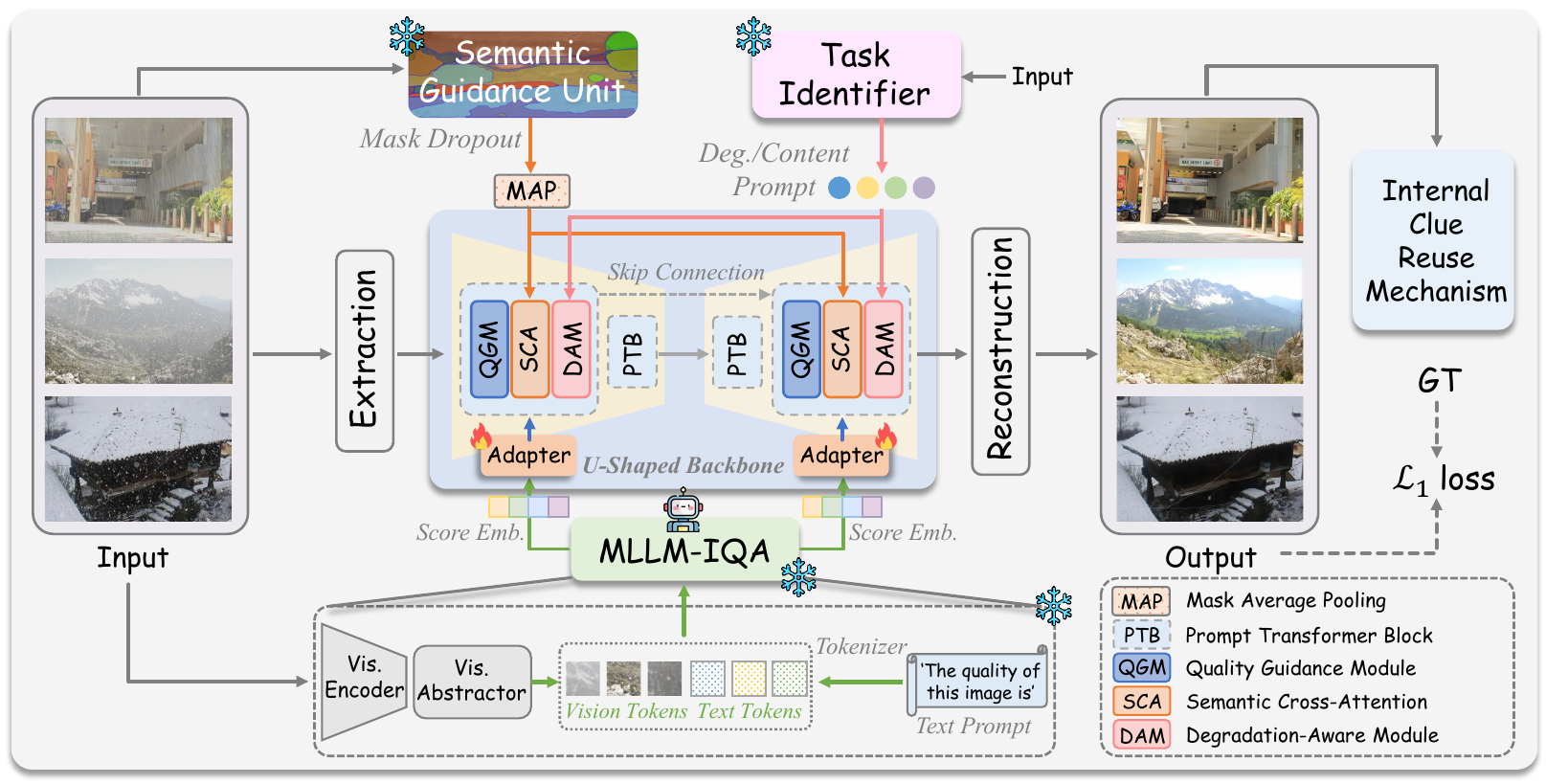}
\caption{An overview architecture of the proposed ClearAIR.}
\label{fig:model}
\end{figure*}

\section{Related Work}

\subsection{All-in-One Image Restoration} 
All-in-One image restoration \cite{EvoIR, WG_TIP, UniUIR, WG-AIR, wr_AIOVR} has emerged as a promising direction in low-level vision, aiming to restore clean images from diverse degradation types using a single unified model. Compared with task-specific \cite{AFDformer, WR_25TIP, CLIP_UIE, WR_24TIP,xiao2024event, ZY_TBC,ZR_24CVPR,ZY_VCIP,xiao2025event, CLB_25AAAI, CLB_25tmm, xiao2025spiking, xiao2025vsr, KLS, DIBR_ESWA, pl1, YH_IQA} and general restoration methods \cite{yuanbiao_IR1, MDDA, yuanbiao_IR2}, all-in-one approaches offer significant advantages in multi-task capability, making it more suitable for practical applications with diverse degradation scenarios. For example, AirNet \cite{AirNet} introduced a contrastive learning strategy to learn discriminative degradation representations. Prompt-based methods such as PromptIR \cite{PromptIR} and ProRes \cite{ProRes} further improved multi-degradation handling by incorporating vision prompts into the network. More recently, DA-CLIP \cite{DA-CLIP} and MPerceiver \cite{MPerceiver} leveraged pre-trained large-scale vision models to boost performance on complex restoration tasks. 
Perceive-IR \cite{PerceiveIR} showed that jointly recognizing degradation types and severity improves restoration, underscoring the importance of comprehensive degradation perception in all-in-one frameworks.

Despite these advancements, most all-in-one image restoration methods adopt a uniform processing strategy, failing to account for the spatial variability of degradation. Moreover, even in cases of uniformly distributed degradation, the difficulty of restoration varies significantly depending on the texture complexity of different regions. For example, flat regions are generally easier to restore, while areas with complex textures pose greater challenges.

\subsection{Human Visual Perception}
In visual cognition, humans exhibit specific characteristics. Typically, a visual image is first perceived as a unified whole before being analyzed in terms of its constituent parts. Recently, several studies have leveraged this perceptual mechanism to achieve promising results. For instance, Dream \cite{Dream} reversely models the hierarchical processing of the Human Visual Perception (HVP) into a computable encoding-decoding framework, revealing that both biological and artificial vision systems aim for efficient visual information coding at their core. In this paper, we propose a hierarchical image perception pipeline that mimics human visual processing: global quality assessment to semantic-driven regional localization and degradation identification of distorted regions. By integrating global coarse-grained understanding and local fine-grained perception, our method not only improves the visual naturalness of restored images but also maintains semantic consistency under complex degradation, enabling more human-like AiOIR.

\subsection{Image Quality Assessment} % y
% {Image Quality Assessment Application in Image Restoration
Image Quality Assessment (IQA) methods primarily rely on quality scores to evaluate image quality. These methods can be categorized into reference and no-reference approaches, depending on whether they require a high-quality reference image. In the field of image restoration, no-reference IQA is commonly used because it can directly regress a quality score without needing a reference image, aligning well with practical restoration needs.
In recent years, Multi-Modal Large Language Models (MLLM)-based IQA methods have leveraged the foundational knowledge of MLLM to achieve superior performance and more detailed assessment results \cite{Q-bench, Q-instruct}. Q-Bench demonstrates that general MLLM possess some low-level perception capabilities. Q-Instruct further enhances these capabilities by introducing a large-scale dataset.
Recently, DeQA \cite{DeQA} has advanced this area by using MLLM to regress precise quality scores, achieving remarkable performance. In this paper, we aim to harness the powerful ability of MLLMs to mine multi-modal cues, employing them as an initial estimator for overall image quality. This serves as a solid foundation for subsequent human visual perception processes.

% *****************************************************
\begin{figure}[!htb]
 % \vspace{-2mm}
\centering
\includegraphics[width=1\columnwidth]{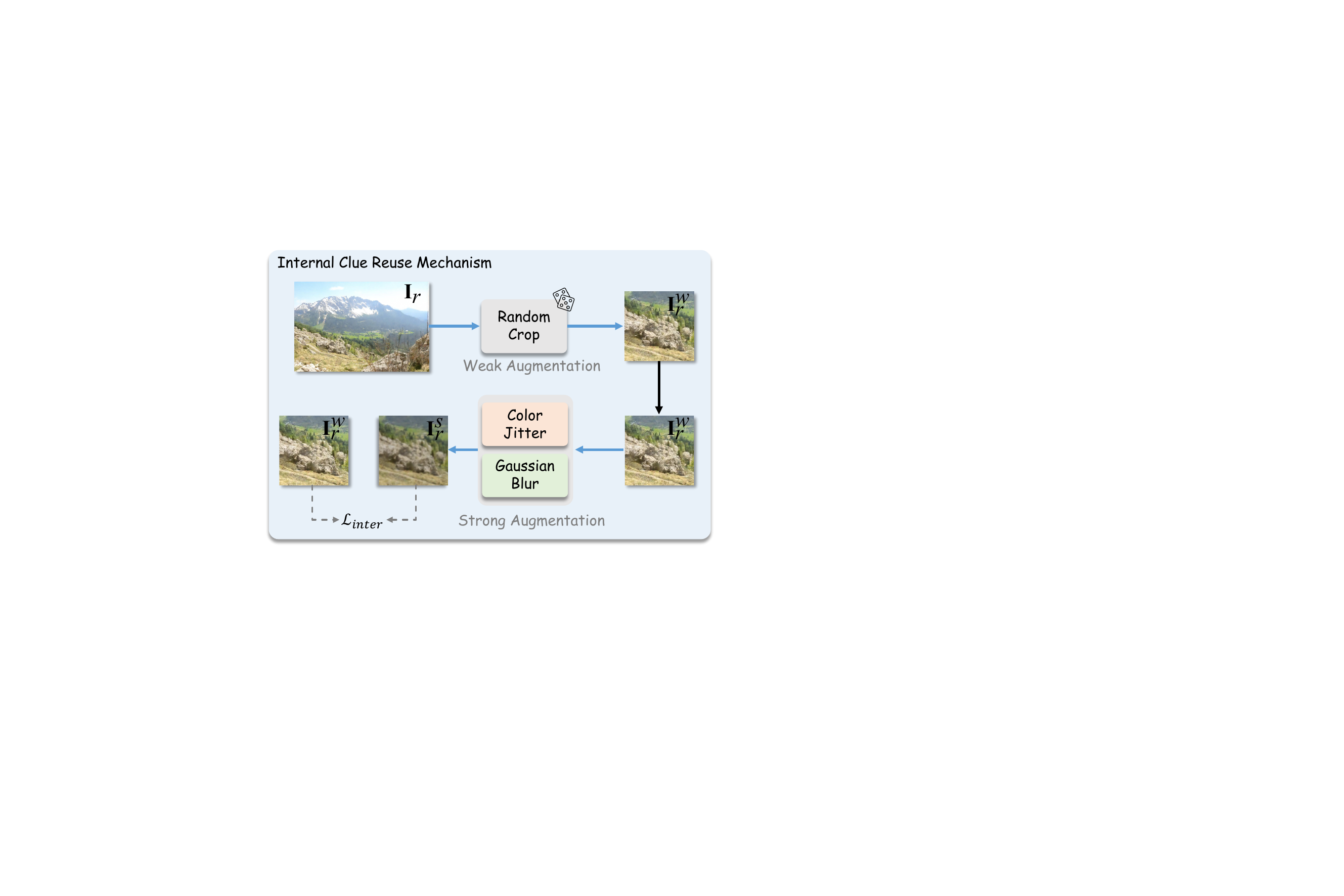}
% \vspace{-1em}
\caption{
The proposed Internal Clue Reuse Mechanism.
}
\label{fig:icrm}
\end{figure}

\section{Methodology}
\label{Methodology}

\subsection{Overall Pipeline} % 缩写了内容

The overall framework is illustrated in Fig.~\ref{fig:model}. ClearAIR consists of four components: 1) MLLM-based IQA: It generates a score embedding from visual and textual tokens, which guides the restoration backbone via the Quality Guidance Module (QGM). 2) Semantic Guidance Unit (SGU): It provides region-level semantic masks, with features fused through Semantic Cross-Attention (SCA). 3) Task Identifier: It estimates the predominant degradation type and encodes this prediction into a degradation prompt used by the Degradation-Aware Module (DAM). 
4) Internal Clue Reuse Mechanism (ICRM): It exploits self-supervised learning to extract internal image cues, enhancing fine-detail reconstruction.
The optimization objective of the overall process can be represented as:
\begin{equation}
    \mathcal{L}_{{total}} = \mathcal{L}_{{1}}+\alpha \cdot \mathcal{L}_{{inter}},
    \label{eq:internal_loss}
\end{equation}
where $\alpha$ is a hyperparameter set to 0.25. 
The details of $\mathcal{L}_{{inter}}$ are provided in the ICRM subsection.

\paragraph{Overall Assessment.} 
Inspired by the early HVP stages, which emphasize global structural cues, we incorporate an MLLM‑IQA model to assess the overall quality of the input image. 
As shown in Fig. 2, a vision encoder is used to encode the input image into visual tokens. In addition, a vision abstractor is utilized as part of the connector module, which further compresses the visual tokens. Finally, the visual and textual tokens are fused and fed into a large MLLM for response prediction.

We extract the state $\mathcal{Q}$ from the layer preceding the `quality level' token. This representation more faithfully captures the MLLM‑IQA model’s underlying reasoning about image quality.
Subsequently, $\mathcal{Q}$ is integrated into the QGM as score embeddings via an affine transformation. Given a degraded image $\mathbf{I}_{d}$ along with its corresponding textual description $\mathcal{T}_{d}$, the above process can be expressed as follows:
\begin{equation}
    \mathcal{Q} = \mathcal{M}_{iqa}(\mathbf{I}_{d}, \mathcal{T}_{d}), 
\end{equation}
\begin{equation}
    \mathbf{F}_q = \mathcal{A}_{adapter}(\mathcal{Q}), 
\end{equation}
\begin{equation}
    \mathbf{X}_{qgm}^{out} = \mathbf{X}_{qgm}^{in}\odot\text{Linear}(\mathbf{F}_q)+\text{Linear}(\mathbf{F}_q),
\end{equation}
where $\mathcal{M}_{iqa}(\cdot)$ refers MLLM-based IQA model (DeQA). $\mathbf{F}_q$ denotes the features after being transformed by the adapter $\mathcal{A}_{adpter}$.
$\mathbf{X}_{qgm}^{in}$ and $\mathbf{X}_{qgm}^{out}$ represent  the input and output features of the QGM, respectively. 
% *********************

\paragraph{Region Awareness.} 
We design a region awareness pipeline to support region‑level segmentation and provide coarse guidance for locating areas that are likely affected by degradation. Specifically,
we introduce the SGU, which leverages a pre-trained Segment Anything Model \cite{SAM2, RF_SAM} to extract high-level semantics. 
Given a degraded image $\mathbf{I}_d$, SGU generates $N_m$ binary masks:
\begin{equation}
\mathbf{I}_{{mask}} \in \{0,1\}^{H \times W \times N_m},
\end{equation}
where each mask $m_i \in \mathbb{R}^{H \times W \times1}$ highlights a distinct region.
These masks are integrated with shallow features $\mathbf{F}_s \in \mathbb{R}^{H \times W \times C}$ via {Mask Average Pooling (MAP)}. For each mask $m_i$, we compute the average feature within the masked region and broadcast it back:
\begin{equation}
\bar{\mathbf{f}}_i = \frac{1}{|\Omega_i|} \sum_{(h,w) \in \Omega_i} \mathbf{F}_s(h,w),
\end{equation}
\begin{equation}
\mathbf{F}_{{sem}}(h,w) = \bar{\mathbf{f}}_i, \; 
\forall (h,w) \in \Omega_i,
\end{equation}
where $\Omega_i = \{(h,w) \mid m_i(h,w) = 1\}$. The output $\mathbf{F}_{{sem}} \in \mathbb{R}^{H \times W \times C}$ encodes semantic-aware structural priors.

To enhance robustness to fluctuations in mask quality resulting from degradation severity or model scale, we introduce mask dropout during training, removing a random subset of masks and merging their regions into the background.
Finally, $\mathbf{F}_{{sem}}$ interacts with the restoration backbone through SCA enabling region-level semantic guidance in the restoration process. This process can be expressed as follows:
\begin{equation}
\mathbf{Q} = \mathbf{F}^{in}_{sca},\ 
\mathbf{K} = \mathbf{W}_k \mathbf{F}_{{sem}},\ 
\mathbf{V} = \mathbf{W}_v \mathbf{F}_{{sem}},
\end{equation}
\begin{equation}
\mathbf{F}^{out}_{sca} = {Softmax}\left(\frac{\mathbf{Q}\mathbf{K}^T}{\sqrt{d}}\right) \mathbf{V},
\end{equation}
where $\mathbf{F}^{in}_{sca}$ and $\mathbf{F}^{out}_{sca}$ represent the input and output features of the SCA module, respectively.

% ********************************************************
\paragraph{Task Recognition.}  
In this part, we primarily predict local degradation types, enabling a more informed characterization of region‑level degradation patterns. Specifically, we employ DA-CLIP \cite{DA-CLIP} as Task Identifier to generate both content embeddings $\mathbf{F}_{c} \in \mathbb{R}^{1 \times 512}$ and degradation embeddings $\mathbf{F}_{d}\in \mathbb{R}^{1 \times 512}$. 
% Given the diversity of degradation types in our experimental setup, we conduct pre-training for each specific degradation pattern. 
The degradation embedding is then transformed into a degradation prompt $\mathbf{F}_{{p}}$, which can be described as:
\begin{equation}
    \mathbf{F}_{p} = MLP\Big(\mathcal{P} \odot {Softmax}\big(MLP(\mathbf{F}_{{d}})\big)\Big), 
\end{equation}
where $\mathcal{P}$ denotes a set of learnable prompts. 
Subsequently, the feature $\mathbf{X}_{dam}^{in}$ and $\mathbf{F}_{c}$ are fed into the DAM, enabling cross‑attention for content‑aware spatial enhancement:
\begin{equation}
    \mathbf{\hat{X}}^{in}_{{dam}} = \mathcal{C}_{1\times1}\big(Norm(\mathbf{X}_{dam}^{in})\big), 
\end{equation}
\begin{equation}
    \mathbf{X}_{{dam}}^{att} = CrossAtt.(\mathbf{F}_{{c}}, \mathbf{\hat{X}}^{in}_{{dam}}). 
\end{equation}

Meanwhile, we generate a degradation mask $\mathbf{M}_{d}\in \mathbb{R}^{1 \times h\times w}$ based on $\mathbf{F}_{p}$, and then use $\mathbf{F}_{p}$ to modulate the features $\mathbf{\hat{X}}^{in}_{{dam}}$.  The process can be described as follows:
\begin{equation}
    \mathbf{M}_{{d}} = Sigmoid\big(MLP(\mathbf{F}_{p})\big), 
\end{equation}
\begin{equation}
    \mathbf{F}_{{m}} = \mathbf{M}_{{d}}\odot \mathbf{\hat{X}}^{in}_{{dam}}, 
\end{equation}
where the modulated feature $\mathbf{F}_{{m}}$ is concatenated with $\mathbf{X}_{{dam}}^{att}$ and subsequently fused to get $\mathbf{X}_{{dam}}^{out}$.

\paragraph{Internal Clue Reuse Mechanism.}
\label{subsec:lcrm} 
As shown in Fig. \ref{fig:icrm}, we introduce ICRM to enhance the model’s ability to preserve fine details in restored images. To achieve this, we apply data augmentation with different strengths to the restored output $\mathbf{I}_{{r}}$.
First, weak augmentation is applied to $\mathbf{I}_{{r}}$, formulated as:
\begin{equation}
    \mathbf{I}_{r}^{w} = \mathcal{F}_{weak}(\mathbf{I}_{{r}}),
\end{equation}
where $\mathcal{F}_{weak}(\cdot)$ denotes a weak augmentation operation,
and $\mathbf{I}_{r}^{w}$ is the output image after applying $\mathcal{F}_{weak}(\cdot)$. 
Subsequently, strong augmentation is performed on the $\mathbf{I}_{r}^{w}$, which can be expressed:
\begin{equation}
    \mathbf{I}_{r}^{s} = \mathcal{F}_{strong}(\mathbf{I}_{r}^{w}),
\end{equation}
where $\mathcal{F}_{strong}(\cdot)$ represents a strong augmentation operation, and
$\mathbf{I}_{r}^{s}$ is the resulting strongly augmented image. 
Finally, we compute the L2 distance between the weakly and strongly augmented results to form an internal consistency:
\begin{equation}
    \mathcal{L}_{{inter}} = \gamma \cdot \| \mathbf{I}_{r}^{w} - \mathbf{I}_{r}^{s} \|_2^2,
    \label{eq:internal_loss}
\end{equation}
where $\gamma$ is a hyperparameter that controls the weight of this loss. In our experiments, we set the initial value of $\gamma$ = 0.05.

% ****************************** 数据集设置
\begin{table}[!tb]
\centering
\small

\setlength{\tabcolsep}{0.5pt} % 调整列宽命令，可以被AAAI允许
% \resizebox{0.95\linewidth}{!} 
{\begin{tabular}{lcc} 

\toprule[1pt]
\toprule[0.5pt]

Settings & No. Datasets  & Degradation Type  
% & Degradation Mode
\\

\midrule
Three Deg.
& 3 
& Noise, Haze, Rain 
% & Separate
\\

Five Deg. 
% \cite{IDR} 	
& 5 
& Noise, Haze, Rain, Blur, Low-light
% & Separate
\\

All-Weather 
% \cite{Transweather} 
& 3 
& Haze, Rain, Raindrop, Snow
% & Separate \& Mixed
\\

Composited Deg.
% \cite{OneRestore} 	
& 1
& {Haze, Rain, Low-light, Snow}
% & Mixed
\\

\bottomrule[1pt]
\end{tabular}}

\caption{
The details of the All-in-One setting.
}
\label{tab:dataset}
% \vspace{-1em}
\end{table}

% ************************************************* 3任务
\begin{table*}[!htb]
    \centering
    \small
    % \fboxsep0.75pt % 框内内容与边框之间的距离
    \setlength\tabcolsep{1.5pt} % 单元格之间的水平间距

    \begin{tabular}{lcccccccccccccc}
    
    \toprule[1pt]    
    \toprule[0.5pt]

     \multirow{2}{*}{Method} 
     & \multirow{2}{*}{Source}
     & \multirow{2}{*}{Params.} 
     & \multicolumn{2}{c}{\textit{Dehazing}} 
     & \multicolumn{2}{c}{\textit{Deraining}} 
     & \multicolumn{6}{c}{\textit{Denoising}} 
     & \multicolumn{2}{c}{\multirow{2}{*}{Average}}  
     \\
     
     \cmidrule(lr){4-5} \cmidrule(lr){6-7} \cmidrule(lr){8-13} 
     &
     &
     & \multicolumn{2}{c}{SOTS} 
     & \multicolumn{2}{c}{Rain100L} 
     & \multicolumn{2}{c}{BSD68\textsubscript{$\sigma$=15}} 
     & \multicolumn{2}{c}{BSD68\textsubscript{$\sigma$=25}} 
     & \multicolumn{2}{c}{BSD68\textsubscript{$\sigma$=50}} 
     &  
     \\
     
     \midrule
        DL \cite{DL}
        & TPAMI'19
        & 2M 
        & 26.92 & {.931} 
        & 32.62 & {.931} 
        & 33.05 & {.914} 
        & 30.41 & {.861} 
        & 26.90 & {.740}  
        & 29.98 & {.876}
        \\
        
        AirNet \cite{AirNet}
        & CVPR'22
        & 9M 
        & 27.94 &  {.962} 
        & 34.90 &  {.967} 
        & 33.92 & {{{.933}}} 
        & 31.26 &  {{{.888}}} 
        & 28.00 &  {{{.797}}} 
        & 31.20 &  {.910} 
        \\

        IDR \cite{IDR}
        & CVPR'23
        & 15M 
        & 29.87 &  {.970} 
        & 36.03 &  {.971} 
        & 33.89 & {{{.931}}} 
        & 31.32 &  {{{.884}}} 
        & 28.04 &  {{{.798}}} 
        & 31.83 &  {.911} 
        \\
        
        PromptIR \cite{PromptIR}
        & NeurIPS'23
        & 36M 
        & {{30.58}} &  {{{.974}}} 
        & {{36.37}} &  {{{.972}}} 
        & {{33.98}} &  {{{.933}}} 
        & {{31.31}} &  {{{.888}}} 
        & {{28.06}} &  {{{.799}}} 
        & {{32.06}} &  {{{.913}}} 
        \\

        NDR \cite{NDR}
        & TIP'24
        & 28M 
        & 28.64 &  {.962} 
        & 35.42 &  {.969} 
        & 34.01 & {{{.932}}} 
        & 31.36 &  {{{.887}}} 
        & 28.10 &  {{{.798}}} 
        & 31.51 &  {.910} 
        \\

        Gridformer \cite{Gridformer}
        & IJCV'24 
        & 34M 
        & 30.37 &  {.970} 
        & 37.15 &  {.972} 
        & 33.93 & {{{.931}}} 
        & 31.37 &  {{{.887}}} 
        & 28.11 &  {{{.801}}} 
        & 32.19 &  {.912} 
        \\

        InstructIR \cite{InstructIR} 
        & ECCV'24 
        & 16M 
        & 30.22 &  {.959} 
        & 37.98 &  {.978} 
        & \underline{34.15} & {{{.933}}} 
        & {\underline{31.52}} & {{{.890}}} 
        & {\underline{28.30}} &  {\underline{{.803}}}
        & 32.43 &  {.913} 
        \\

        Perceive-IR \cite{PerceiveIR}
        & TIP'25 
        & 42M 
        & 30.87 &  {.975} 
        & 38.29 &  {.980} 
        & {34.13} & \underline{.934}
        & \bf{31.53} & {.890} 
        & {\textbf{28.31}} &  {\textbf{{.804}}}
        & 32.63 &  {.917} 
        \\

        AdaIR \cite{AdaIR} 
        & ICLR'25
        & 29M 
        & \underline{31.06} &  \underline{.980} 
        & \underline{38.64} &  \underline{.983} 
        & 34.12 & \underline{.934}
        & 31.45 &  {\textbf{{.892}}}
        & 28.19 &  {{{.802}}} 
        & 32.69 &  {\underline{{.918}}} 
        \\

        VLU-Net \cite{VLU-Net} 
        & CVPR'25
        & 35M 
        & 30.71 &  \underline{.980} 
        & {\bf38.93} &  {\bf.984} 
        & 34.13 & {\textbf{{.935}}} 
        & 31.48 &  {\textbf{{.892}}} 
        & 28.23 &  {\textbf{{.804}}} 
        & {\underline{32.70}} &  {\textbf{{.919}}} 
        \\
    
        % \midrule
        
        {ClearAIR} (Ours) 
        & -
        & {{31M}} 
        & {\textbf{31.08}} &  {\textbf{{.981}}} 
        & {{38.61}} &  {\textbf{{.984}}} 
        & {\textbf{34.18}} &  {\bf.935} 
        & {{31.50}} &  \underline{.891} 
        & {\textbf{28.31}} &  {\textbf{{.804}}} 
        & {\textbf{32.74}} &  {\textbf{{.919}}} 
        \\

     \bottomrule[1pt]
    \end{tabular}
    \caption{Comparison to state-of-the-art AiOIR methods on the Three Degradations task.}
    \label{tab:3deg}  
\end{table*}

% *************************************************3_degradation
\begin{figure*}[!htb]
\centering
\includegraphics[width=0.87\textwidth]{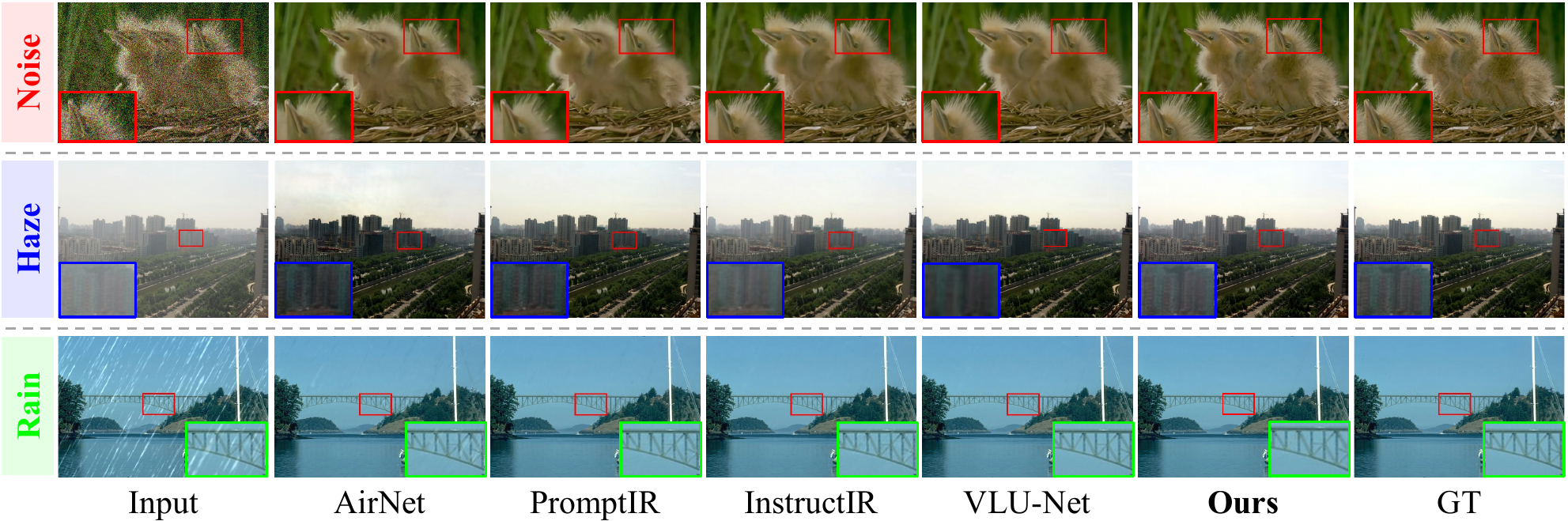}
\caption{Visual comparisons of ClearAIR with state-of-the-art AiOIR methods on the Three Degradations task.}
\label{fig:3_degs}
\end{figure*}

% ******************************************************************
\section{Experiments}
\label{sec:exp}

\subsection{Experimental Setup}
\label{experiments}

\paragraph{Datasets.} % y
We conduct experiments under two settings: All-in-One and Single-task following the protocols established in prior works \cite{PerceiveIR}. As shown in Tab. \ref{tab:dataset}, the All-in-One setting entails a comprehensive evaluation across four tasks, with further details on the Single-task setting provided in the \textbf{Appendix}.

% ******************************************************************
\paragraph{Implementation Details.} 
We employ DeQA \cite{DeQA} as the MLLM-IQA model and chose Restormer \cite{Restormer} as the restoration backbone. 
Specifically, from level-1 to level-4, the numbers of Prompt Transformer Block (PTB) are set to [3, 5, 6, 8], the attention heads are [1, 2, 4, 8], and the channel dimensions are [48, 96, 192, 384]. We optimize the network using AdamW ($\beta_1 = 0.9$, $\beta_2 = 0.999$) with a learning rate of $2 \times 10^{-4}$ and batch size of 4. Training runs for 300K iterations. The total loss weights are set as $\lambda_1 = 0.1$ and $\lambda_2 = 0.05$.
All experiments are conducted on NVIDIA GeForce RTX 4090 GPUs. During training, inputs are randomly cropped to 256$\times$256 patches, and random horizontal and vertical flips are applied for data augmentation.

% ***************************************************
\subsection{All-in-One Image Restoration Results}
\label{All-in-One Image Restoration Results}

\paragraph{Three Degradations Task.} % aaai精简版本

We evaluate our model on three restoration tasks: denoising, dehazing, and deraining. 
Tab. \ref{tab:3deg} shows ClearAIR achieves the best average performance, with notable improvements in high-noise removal and severe haze reduction. Outperforming or matching AdaIR and VLU-Net via better human visual perception modeling, it reaches 31.08 dB PSNR on SOTS (vs. 30.71 dB of VLU-Net), proving perceptual enhancement offsets the lack of physical priors. Qualitative results (Fig. \ref{fig:3_degs}) further verify its efficacy in texture preservation, rain streak removal with sharpness retention, and contrast/detail restoration in dense haze.

% *************************************************
\paragraph{Five Degradations Task.} % y

% ************************************************* 5任务
\begin{table*}[!htb]
    \centering
    \small
    % \fboxsep0.75pt
    \setlength\tabcolsep{1.2pt}

    \begin{tabular}{lcccccccccccccc}
    
    \toprule[1pt]
    \toprule[0.5pt]
    
     \multirow{2}{*}{Method} 
     & \multirow{2}{*}{Source}
     & \multirow{2}{*}{Params.} 
     & \multicolumn{2}{c}{\textit{Dehazing}} 
     & \multicolumn{2}{c}{\textit{Deraining}} 
     & \multicolumn{2}{c}{\textit{Denoising}} 
     & \multicolumn{2}{c}{\textit{Deblurring}} 
     & \multicolumn{2}{c}{\textit{Low-Light}} 
     & \multicolumn{2}{c}{\multirow{2}{*}{Average}}  
     \\
     
     \cmidrule(lr){4-5} 
     \cmidrule(lr){6-7} 
     \cmidrule(lr){8-9} 
     \cmidrule(lr){10-11} 
     \cmidrule(lr){12-13}
     
     &
     & 
     &
     \multicolumn{2}{c}{SOTS} 
     & \multicolumn{2}{c}{Rain100L} 
     & \multicolumn{2}{c}{BSD68\textsubscript{$\sigma$=25}} 
     & \multicolumn{2}{c}{GoPro} 
     & \multicolumn{2}{c}{LOLv1} 
     &  
     \\
     
     \midrule

    DL \cite{DL}
    & TPAMI'19
    & 2M 
    & 20.54 
    &  {.826} 
    & 21.96 
    &  {.762} 
    & 23.09 
    &  {.745} 
    & 19.86 
    &  {.672} 
    & 19.83 
    &  {.712} 
    & 21.05 
    &  {.743} 
    \\

    AirNet \cite{AirNet}
    & CVPR'22 
    & 9M 
    & 21.04 &  {.884} 
    & 32.98 &  {.951} 
    & 30.91 &  {{{.882}}} 
    & 24.35 &  {.781} 
    & 18.18 &  {.735} 
    & 25.49 &  {.847} 
    \\

      IDR \cite{IDR}
      & CVPR'23 
      & 15M 
      & {{25.24}} &  {{{.943}}} 
      & {{35.63}} &  {{{.965}}} 
      & {{\bf31.60}} &  {{{.887}}} 
      & {{27.87}} &  {{{.846}}} 
      & {{21.34}} &  {{{.826}}} 
      & {{28.34}} &  {{{.893}}} 
      \\

     PromptIR \cite{PromptIR}
     & NeurIPS'23 
     &  {{33M}} 
     &  {{26.54}} &  {{.949}}
     &  {{36.37}} &   .970
     &  {{31.47}} &  {.886} 
     &  {{28.71}} &   {.881}
     &  22.68     &   .832
     &  {29.15}   &   {.904}
     \\

     Gridformer \cite{Gridformer}
     & IJCV'24 
     &  {{34M}} 
     &  {{26.79}} &  {{.951}}
     &  {{36.61}} &   .971
     &  {{31.45}} &  {.885} 
     &  {{29.22}} &   {.884}
     & 22.59      &   .831
     &  {29.33} &   {.904}
     \\

     InstructIR \cite{InstructIR}
     & ECCV'24  
     &  {{16M}} 
     &  {{27.10}} &  {{.956}}
     &  {{36.84}} &   .973
     &  {{31.40}} &  {.887} 
     &  29.40 &   \underline{.886}
     &  {\bf23.00}     &   .836
     &  {29.55}   &   {.907}
     \\

    Perceive-IR \cite{PerceiveIR}
    & TIP'25 
    & 42M 
    & 28.19 &  {.964} 
    & 37.25 &  {.977} 
    & 31.44 & {.887}
    & \underline{29.46} & \underline{.886}
    & 22.81 &  {{{.833}}}
    & 29.84 &  {.909} 
    \\

     AdaIR \cite{AdaIR}
     & ICLR'25 
     &  {29M}
     &  \underline{30.53} &  \underline{.978}
     &  {{38.02}} &   \underline{.981} 
     &  {{31.35}} &  \underline{.888} 
     &  {{28.12}} &   {{{.858}}} 
     & {\textbf{23.00}} &   \underline{.845} 
     &  \underline{30.20} &   \underline{.910}
     \\

     VLU-Net \cite{VLU-Net}
     & CVPR'25   
     &  {{35M}} 
     &  {\textbf{30.84}} &  {\textbf{{.980}}} 
     &  {\textbf{38.54}} &   {\textbf{{.982}}} 
     &  {{31.43}} &  {{\bf.891}}  
     &  {{27.46}} &   {.840}
     & 22.29      &   .833
     &  {30.11} &   {.905}
     \\

    % \midrule

     {ClearAIR} (Ours)  
     & -
     &  {{31M}} 
     &  {{30.12}} &  \underline{.978} 
     &  \underline{38.20} &   \textbf{.982} 
     &  \underline{31.53} &  \underline{.888} 
     &  {\textbf{29.67}} &   {\textbf{{.887}}} 
     & \underline{22.83} &   {\textbf{{.846}}} 
     &  {\textbf{30.45}} &   {\textbf{{.916}}} 
     \\

     \bottomrule[1pt]
    \end{tabular}
    \caption{{Comparison to state-of-the-art AiOIR methods on the Five Degradations task.}}
    \label{tab:5deg}
    % \vspace{-0.5em}
\end{table*}

% ********************************************* All-Weather
\begin{table*}[!htb]
    \centering
    \small
    \setlength\tabcolsep{0.8pt}

    \begin{tabular}{lcccccccccccc}

    \toprule[1pt]
    \toprule[0.5pt]
    
     \multirow{2}{*}{Method} 
     & \multirow{2}{*}{Source} 
     & \multirow{2}{*}{Params.} 
     & \multicolumn{2}{c}{\textit{Snow100K-S}} 
     & \multicolumn{2}{c}{\textit{Snow100K-L}} 
     & \multicolumn{2}{c}{\textit{Outdoor-Rain}} 
     & \multicolumn{2}{c}{\textit{RainDrop}} 
     & \multicolumn{2}{c}{{Average}}  
     \\
     
     \cmidrule(lr){4-5} 
     \cmidrule(lr){6-7} 
     \cmidrule(lr){8-9} 
     \cmidrule(lr){10-11} 
     \cmidrule(lr){12-13}
     
     &
     &
     & PSNR
     & SSIM
     & PSNR
     & SSIM
     & PSNR
     & SSIM
     & PSNR
     & SSIM
     & PSNR
     & SSIM
     
     \\
     \midrule

    All-in-One \cite{All_in_one}
    & CVPR'20
    & - 
    & - &  {-} 
    & 28.33 &  {.882} 
    & 24.71 &  {.898} 
    & 31.12 &  {.927} 
    & 28.05 &  {.902} 
    \\
    
    Transweather \cite{Transweather}
    & CVPR'22 
    & 38M 
    & 32.51 &  {.934} 
    & 29.31 &  {.888} 
    & 28.83 &  {.900} 
    & 30.17 &  {.916} 
    & 30.20 &  {.909} 
    \\

    TKL \cite{TKL}
    & CVPR'22
    & 29M 
    & 34.42 &  {.947} 
    & 30.22 &  {.907} 
    & 29.27 &  {.915} 
    & 31.81 &  {.931} 
    & 31.43 &  {.925} 
    \\
    
    WGWSNet \cite{WGWS_Net}
    & CVPR'23 
    & 26M 
    & 34.31 &  {.946} 
    & 30.16 &  {.901} 
    & 29.32 &  {{{.921}}} 
    & 32.38 &  {.938} 
    & 31.54 &  {.926} 
    \\

    WeatherDiff \cite{WeatherDiff}
    & TPAMI'23
    & 83M 
    & 35.83 &  {{{.957}}} 
    & {{30.09}} &  {{{.904}}} 
    & 29.64 &  {{{.931}}} 
    & {{30.71}} &  {{{.931}}} 
    & {{31.57}} &  {{{.931}}} 
    \\

     AWRCP \cite{AWRCP}
     & ICCV'23
     &  -
     &  36.92 &  {.965}
     &  31.92 &  .934
     &  31.39 &  {.933} 
     &  31.93 &  {.931}
     &  33.04 &  .941
     \\

     Histoformer \cite{Histoformer}
     & ECCV'24
     &  {30M}
     &  {{37.41}} &  \underline{.966}
     &  {{32.16}} &   {{.926}}
     &  \underline{32.08} &  \underline{.939} 
     &  {{\bf33.06}} &   {{{\bf.944}}} 
     & \underline{33.68} &   \underline{.945} 
     \\

     T$^{3}$-DiffWeather \cite{T3DiffWeather}
     & ECCV'24
     &  69M
     &  \underline{37.51} &  \underline{.966}
     &  \underline{32.37} &   {\textbf{{.936}}} 
     &  {{31.09}} &  {{.937}} 
     &  {{32.66}} &   {{{.941}}} 
     & {{33.41}} &   \underline{.945} 
     \\

    % \midrule

     {ClearAIR} (Ours)  
     & -
     &  {{31M}} 
     &  {{\bf37.79}} &  {{\textbf{{.967}}}} 
     &  {\textbf{32.53}} &   \underline{.932} 
     &  {\textbf{32.45}} &  {{\bf.941}} 
     &  \underline{32.82} &   \underline{.942} 
     & {\textbf{33.90}} &   {\textbf{{.946}}} 
     \\

     \bottomrule[1pt]
    \end{tabular}
    \caption{{Comparison to state-of-the-art All-in-One methods on the All-Weather task.}}
    \label{tab:weather_removal}
\end{table*}

% ******************************************* 多天气
\begin{figure*}[!htb]
\centering
\includegraphics[width=0.9\textwidth]{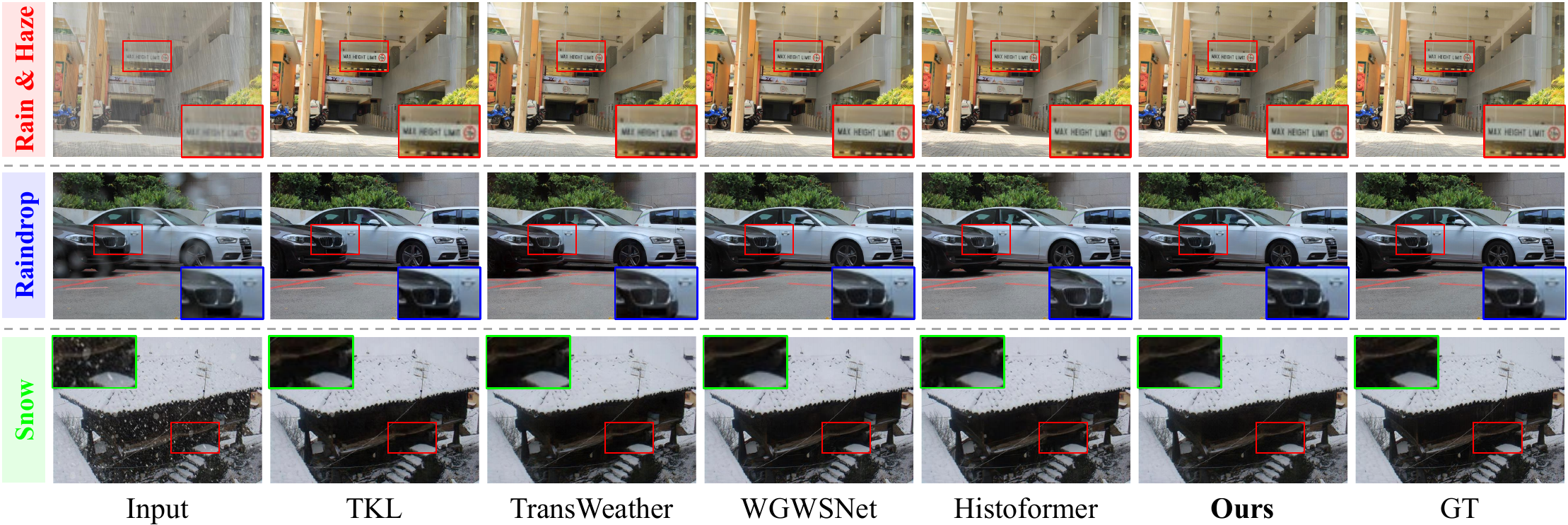} 
% \vspace{-2mm}
\caption{
Visual comparisons of ClearAIR with state-of-the-art AiOIR methods on the All-Weather task. 
} 
\label{fig:weather_deg}
\end{figure*}

We extend ClearAIR to five degradation tasks, using GoPro for deblurring and LOL for low-light enhancement. As shown in Tab.~\ref{tab:5deg}, ClearAIR achieves superior performance on most tasks, excelling particularly in deblurring. Although slightly behind specialized methods in low-light enhancement and denoising, it remains highly competitive and achieves the highest average PSNR (30.45 dB) and SSIM (0.916), demonstrating strong multi-task capability. Additional visual results are provided in the \textbf{Appendix}.

% *********************************************** 混合退化
\begin{figure*}[!htb]
\centering
\includegraphics[width=0.91\textwidth]{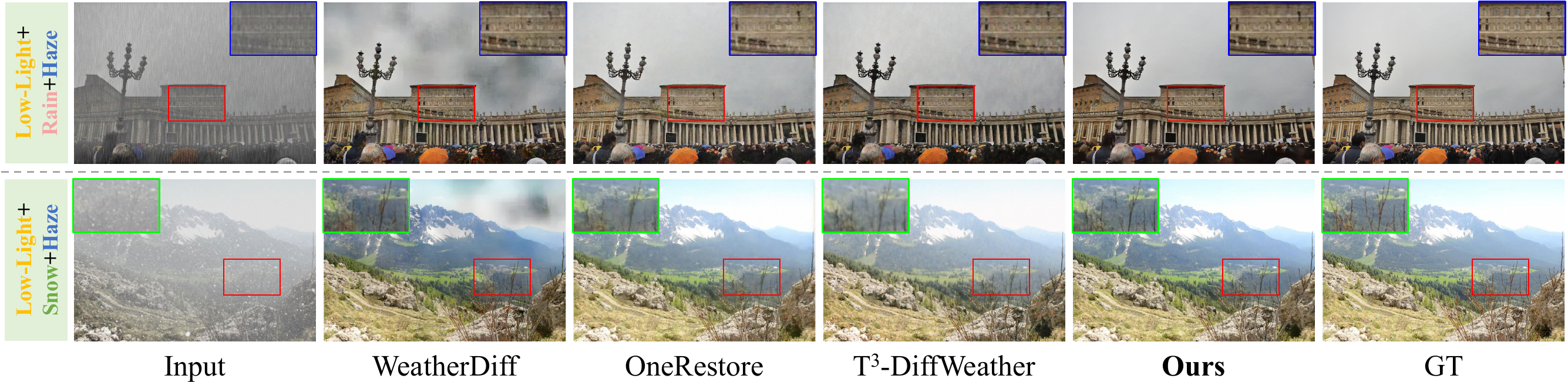}
% \vspace{-2mm}
\caption{
Visual comparisons of ClearAIR with state-of-the-art AiOIR methods on the Composited Degradation task. 
}
% \vspace{-4mm}
\label{fig:cdd11}
\end{figure*}

% ************************************************
\paragraph{All-Weather Task.} % y

We evaluate ClearAIR on All-Weather task including: snow, rain \& haze, and raindrop. 
Tab. \ref{tab:weather_removal} shows ClearAIR achieves an average gain of 0.22 dB over
Histoformer \cite{Histoformer}. These consistent gains across diverse weather conditions demonstrate its effectiveness in handling complex adverse weather degradations. Qualitative comparisons in Fig.~\ref{fig:weather_deg} further illustrate this: ClearAIR produces clearer, more natural results, effectively removing weather artifacts while better preserving details and textures.

% ***********************************************************
\paragraph{Composited Degradation Task.} % y

We evaluate ClearAIR under challenging CDD-11 dataset \cite{OneRestore}, considering both individual and combined degradation.
As shown in Tab.~\ref{tab:CDD11}, ClearAIR achieves an average gain of 0.62 dB over OneRestore, outperforming existing All-in-One models, which validates its effectiveness in modeling human visual perception. Qualitative results in Fig.~\ref{fig:cdd11} demonstrate superior removal of composite degradations while preserving fine details and textures.

% *********************************************** CDD11
\begin{table}[!htb]
\centering
\small

\setlength{\tabcolsep}{5pt}
\begin{tabular}{lcccc}

\toprule[1pt]
\toprule[0.5pt]

Method 
& Source
& Params.
& PSNR$\uparrow$ 
& SSIM$\uparrow$
\\

\midrule
AirNet
& CVPR'22
& 9M
& 23.75 
& 0.814 
\\   

TransWeather
& CVPR'22
& 38M
& 23.13 
& 0.781
\\   

WeatherDiff
& TPAMI'23
& 83M
& 22.49 
& 0.799 
\\

PromptIR
& NeurIPS'23
& 33M
& 25.90 
& 0.850 
\\   

WGWSNet
& CVPR'23
& 26M
& 26.96 
& 0.863 
\\   

OneRestore
& ECCV'24
& 6M
& 28.72 
& 0.882 
\\   

% \midrule
{ClearAIR} (Ours)	
& -
& 31M
&	\bf29.34 
&	\bf0.886 
\\  

\bottomrule[1pt]
\end{tabular}

\caption{
Comparison to state-of-the-art All-in-One methods on the Composited Degradations task (CDD-11 dataset).
}
% \vspace{-1em}
\label{tab:CDD11}
\end{table}

% *********************************
\subsection{Ablation Study}

This section analyzes the impact of different design choices in ClearAIR on model performance. All experiments are conducted on the Rain100L dataset \cite{Rain100L} using a training of 100K iterations.

% *********************************
\paragraph{Effects of Perception Order.} % 用3任务的平均

To investigate the impact of perception order, the sequence of overall assessment (How), region awareness and task Recognition (Where and What), we design two additional experimental setups based on the baseline order: Where-What-How and What-How-Where (indexes a and b).  
As shown in Tab.~\ref{tab:ab-perception_order}, the Where-What-How order yields the worst performance. 
This may be because perceiving region-level semantic information first disrupts the structural integrity that is crucial for coarse quality assessment. Notably, the What-How-Where order achieves the second-best result, which aligns with the workflow of some AiOIR methods that begin with degradation characterization. In contrast, our method, inspired by the human visual perception process, achieves the best overall performance.

% ********************************************************************
\begin{table}[!htb]
\centering
\small

{\begin{tabular}{cccc}  
% AAAI禁止使用\resizebox命令 ！！！ ***********************************************

\toprule[1pt]
\toprule[0.5pt]
Index 
& Order 
& PSNR$\uparrow$ 
& SSIM$\uparrow$
\\

\midrule
 
a
& Where-What-How
& 37.89 
& 0.982
\\   

b
& What-How-Where
& 38.04 
& 0.983 
\\

Ours
& How-Where-What 
& 38.21
& 0.986
\\  

\bottomrule[1pt]
\end{tabular}}

\caption{
Effectiveness of perception order.
}
% \vspace{-1em}
\label{tab:ab-perception_order}
\end{table}

% Task Identifier
% ************************************** different components
\begin{table}[!htb] 
\small
\centering
    {\begin{tabular}{ccccccc} % \resizebox{\linewidth}{!}{
  
    \toprule[1pt]
    \toprule[0.5pt]
  
    Index
    &  {IQA} 
    &  SGU 
    &  TI  
    &  ICRM  
    &  PSNR$\uparrow$
    &  SSIM$\uparrow$
    \\

    \midrule

    % \cdashline{1-7}
    a
    & {\ding{51}}  
    & {\ding{55}}  
    & {\ding{55}} 
    & {\ding{51}}
    & 37.57 & 0.980  
    \\

    % \hdashline
    b
&{\ding{55}} 
    & {\ding{51}} 
    & {\ding{55}} 
    & {\ding{51}}
    & 37.43 & 0.978  
    \\

    c
&{\ding{55}} 
    & {\ding{55}} 
    & {\ding{51}}  
    & {\ding{51}}
    & 37.52 & 0.980  
    \\

    d
&{\ding{51}} 
    & {\ding{51}} 
    & {\ding{55}}
    & {\ding{51}}
    & 38.05 & 0.985  
    \\

    e
&{\ding{51}} 
    &{\ding{55}} 
    &{\ding{51}}
    & {\ding{51}}
    & 37.93 & 0.984  
    \\

    f
&{\ding{55}} 
    & {\ding{51}} 
    &{\ding{51}} 
    & {\ding{51}}
    & 37.87
    & 0.984  
    \\

    g
&{\ding{51}} 
    & {\ding{51}} 
    &{\ding{51}} 
    & {\ding{55}}
    & 38.03
    & 0.985  
    \\

        Ours
&{\ding{51}} 
    & {\ding{51}} 
    & {\ding{51}} 
    & {\ding{51}}
    & \bf 38.21 & \bf 0.986  
    \\

  \bottomrule[1pt]
  \end{tabular}}

\caption{Effectiveness of different components.}
  \label{tab:ablation_components} 
  % \vspace{-1em}
\end{table}

% *******************************************
\subsection{Effects of Different Components}

As shown in Tab.~\ref{tab:ablation_components}, we conduct ablation studies to evaluate the contribution of each proposed component. The experimental settings are as follows:
(1) w/o MLLM-IQA (IQA): replaces quality guidance with a learnable parameter;
(2) w/o SGU: replaces semantic priors with learnable parameters;
(3) w/o Task identifier (TI): removes degradation prompts, using a learnable parameter instead;
(4) w/o LCRM: removes the Internal Clue Reuse Mechanism.
Variants (a–c), which replace structured priors with unstructured learnable parameters, exhibit clear performance degradation, demonstrating the necessity of explicit prior modeling for reliable guidance and task adaptation.
Variants (d–f), which preserve partial structured cues (e.g., quality or degradation estimation), yield noticeable improvements, underscoring the value of perceptual and degradation awareness.
Removing LCRM (g) also reduces performance, though to a lesser extent, indicating its benefit in leveraging internal structures and contextual information.

% *************************************************************
\section{Conclusion}

In this paper, we propose ClearAIR, a novel AiOIR framework inspired by HVP and designed with a hierarchical, coarse-to-fine restoration strategy.
By mimicking the HVP’s tendency to first perceive an image as a whole before focusing on local details, our method integrates overall assessment, region awareness, task recognition, and internal clue reuse mechanism for fine-grained restoration. The combination of an MLLM-based image quality assessment model, the semantic guidance unit, and a task identifier enables accurate localization and understanding of degradation patterns. Furthermore, the proposed internal clue reuse mechanism enhances the model's ability to recover detailed textures in a self-supervised manner. Experimental results demonstrate that ClearAIR achieves state-of-the-art performance on both synthetic and real-world datasets.

\section*{Acknowledgments} 
This work was supported by the National Natural Science Foundation of China under Grant 62431020, the National Key Research and Development Program of China under Grant 2024YFE0111800, and the Fundamental Research Funds for the Central Universities under Grant 2042025kf0030.

\bibliography{aaai2026}

\clearpage
\appendix

\twocolumn[
  \begin{center}
    {\LARGE \textbf{Supplementary Material}} % 这里是大标题，\Huge 控制大小
    \vspace{1cm}
  \end{center}
]

% ----------- Supplementary Content Starts Here -----------

\section{Appendix}

\begin{table*}[!htb] % 

  \centering
  \small
  \renewcommand\arraystretch{1.1}
  %   \footnotesize  
    % \vspace{0.5em}
    
  % \tabcolsep=0.2cm  % 列间距命令；和\setlength{\tabcolsep}{1pt} 功能重复了

    {\begin{tabular}{lccccc} % \resizebox{\linewidth}{!}

    \toprule[1pt]
        \toprule[0.5pt]

    \multirow{2}{*}{\bf Method} 
    & \multicolumn{3}{c}{\textbf{RainDS}}
    & \multicolumn{2}{c}{\textbf{Snow100K-real}}  
    \\
    \cmidrule(r){2-4} \cmidrule(r){5-6}

    {} 
    & PSNR$\uparrow$ & SSIM$\uparrow$ & MUSIQ$\uparrow$     
    & MUSIQ$\uparrow$ & NIQE$\downarrow$
    \\

    \midrule

    AirNet \cite{AirNet}
    & 19.67  & 0.594  & 48.42
    & 59.26  & 3.149  
    \\

    TransWeather \cite{Transweather}
    & 21.25  & 0.634  & 48.95
    & 58.79  & 3.036  
    \\

    WeatherDiff \cite{WeatherDiff}
    & 22.62  & 0.659  & \bf55.04
    & 60.47  & 2.951  
    \\

    PromptIR \cite{PromptIR}  
    & 22.83  & 0.654  & 53.17
    & 60.96  & 2.949  
    \\

    WGWSNet \cite{WGWS_Net}
    & \underline{22.97}  & 0.651  & 47.18
    & 61.03  & 2.914  
    \\

    OneRestore \cite{OneRestore}
    & 22.74  & \underline{0.667}  & 53.91
    & \underline{61.24}  & \underline{2.887}  
    \\

\bf{ClearAIR} 
    & \textbf{23.24}  & \textbf{0.675}  & \underline{54.38}    
    & \textbf{61.57}  & \textbf{2.852}  
    \\

  \bottomrule[1pt]
  
  \end{tabular}}
  \caption{Quantitative Results in Real-world All-Weather task.} 
  \label{tab:real_deweather}
  
\end{table*}

\subsection{A. More Details on Datasets and Evaluation}
\paragraph{Datasets.}
Following \cite{PerceiveIR, IDR}, we categorize training setups into ``All-in-One'' and ``Single-task'' based on whether datasets are combined for mixed training. For AiOIR, we train on
a mixed dataset containing multiple degradations and test one
by one on the dataset containing a single type of degradation.
Four All-in-One settings are summarized as follows, each specifying the included degradation types and corresponding datasets:
\begin{itemize}
\item \textbf{Three Degradations:} Gaussian noise (using BSD400 \cite{BSD400}, WED \cite{WED}, and BSD68 \cite{BSD68}), haze (SOTS \cite{SOTS}), and rain (Rain100L \cite{Rain100L}).
    \item \textbf{Five Degradations:} Extends the Three Degradations setting by adding Motion Blur (GoPro \cite{GoPro}) and Low-light (LOL \cite{LOL}).
    \item \textbf{All-Weather:} Focuses on adverse weather conditions: Haze \& Rain (Outdoor-Rain \cite{Outdoor_Rain}), Raindrop (Raindrop \cite{Raindrop}), and Snow (Snow100K \cite{Snow100K}).
    \item \textbf{Composited Degradations:} Challenging combinations of Haze, Rain, Low-light, and Snow are synthesized on the DIV2K dataset \cite{DIV2K} and are known as the CDD-11 benchmark \cite{OneRestore}.
\end{itemize}

Under the ``Single-task'' setting, the model is trained and tested on one specific restoration task at a time. The datasets used for each task are detailed below:
\begin{itemize}
\item \textbf{Image Denoising:} Training is performed on a merged dataset of BSD400 (400 images) and WED (4,744 images). Noisy images are generated by adding Gaussian noise with levels $ \sigma \in \{15, 25, 50\} $. Testing is conducted on the BSD68, Urban100 \cite{Urban100}, and Kodak24 \cite{Kodak24} datasets.
\item \textbf{Image Dehazing:} The OTS subset of RESIDE-$\beta$ \cite{SOTS} (72,135 pairs) is used for training, and the SOTS-Outdoor dataset \cite{SOTS} (500 images) for testing.
\item \textbf{Image Deraining:} The Rain100L dataset is used, with 200 image pairs for training and 100 pairs for testing.
\end{itemize}

\paragraph{Evaluation Metrics.}
% We evaluate performance using reference metrics, including Peak Signal-to-Noise Ratio (PSNR) and Structural Similarity Index (SSIM) \cite{SSIM} as well as non-reference metrics such as the 
% Multi-scale Image Quality Transformer (MUSIQ) \cite{MUSIQ} and Natural Image Quality Evaluator (NIQE) \cite{NIQE}. For the metrics PSNR, SSIM, and MUSIQ, higher scores indicate better performance. In contrast, for the NIQE metric, lower scores are preferred. 

We assess model performance using both reference-based and no-reference image quality metrics. The former includes Peak Signal-to-Noise Ratio (PSNR) and Structural Similarity Index (SSIM) \cite{SSIM}, while the latter comprises the Multi-scale Image Quality Transformer (MUSIQ) \cite{MUSIQ} and the Natural Image Quality Evaluator (NIQE) \cite{NIQE}.
Higher values of PSNR, SSIM, and MUSIQ indicate better quality, whereas lower NIQE scores correspond to superior perceptual quality.

\subsection{B. More Experiment Results}

\paragraph{Generalization to Real-World Scenario.}

\begin{figure*}[!htb]
  % \vspace{-2mm}
\centering
\includegraphics[width=0.9\textwidth]{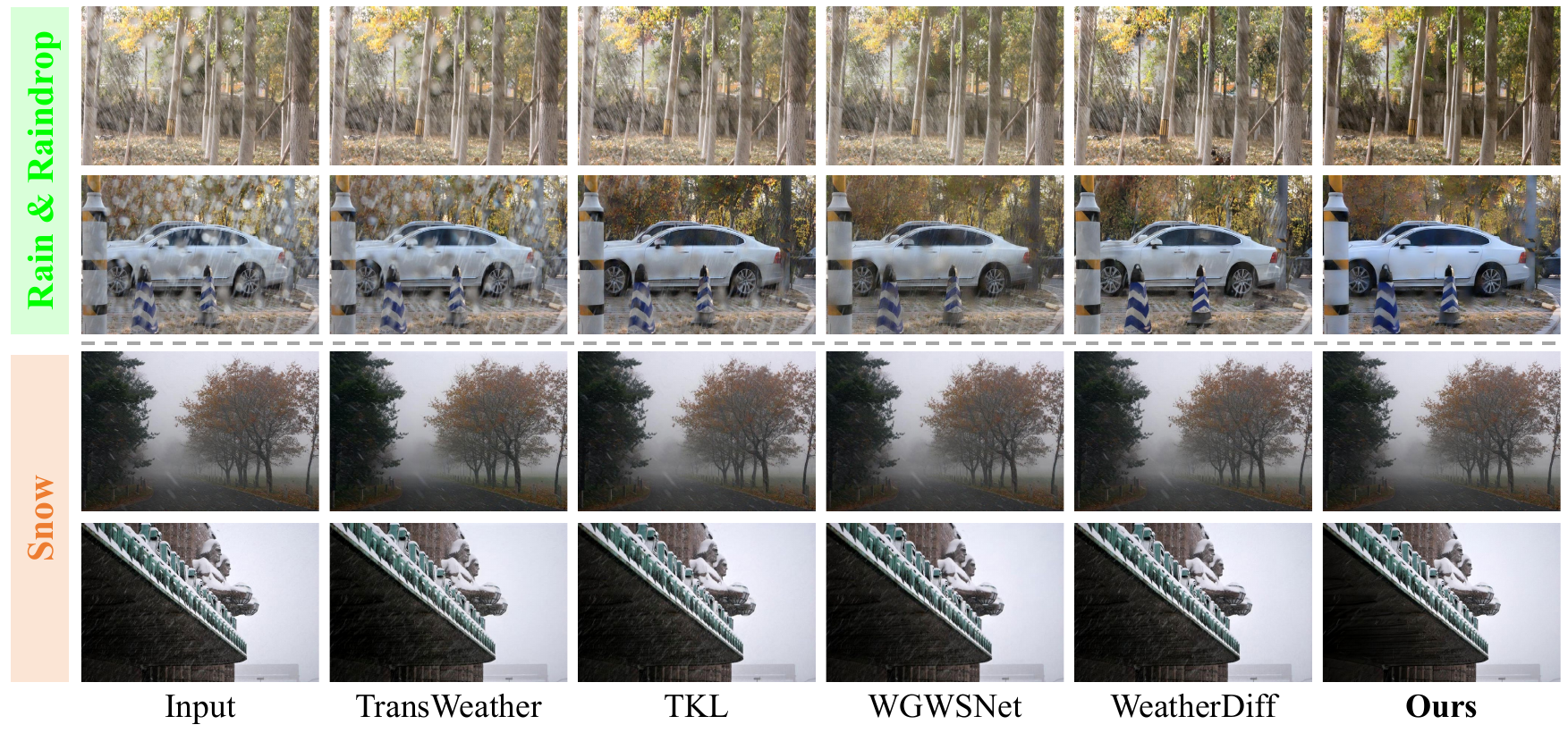}
% \vspace{-4mm}
\caption{
Visual comparisons of {ClearAIR} against state-of-the-art All-in-One methods under real-world All-Weather task on RainDS and Snow100K-real datasets (from top to bottom). Zoom-in for best view.
}
% \vspace{-2mm}
\label{fig:real_weather}
\end{figure*}

We further evaluate the performance of our {ClearAIR} in real-world scenarios under various weather conditions, comparing it with several recent state-of-the-art methods.
The RainDS \cite{RainDS} and Snow100K-real \cite{Snow100K} datasets are used as test benchmarks. All methods are trained on the All-Weather task and directly evaluated on these real-world datasets.
For Snow100K-real, where ground truth images are not available, we employ non-reference image quality assessment metrics, including MUSIQ \cite{MUSIQ} and NIQE \cite{NIQE}, for quantitative evaluation.
As shown in Tab. \ref{tab:real_deweather}, our method achieves the best performance on both RainDS and Snow100K-real datasets.
In Fig. \ref{fig:real_weather}, it is clearly observed that WeatherDiff and WGWSNet fail to effectively suppress weather-related artifacts, while {ClearAIR} produces significantly clearer and more visually pleasing results.
These quantitative and qualitative results demonstrate the superior practical effectiveness of {ClearAIR} in real-world All-Weather task.

\paragraph{Single Degradation Task.}

In this section, we evaluate the performance of ClearAIR under the One-by-One setting. As shown in Tab. \ref{tab:single_denoising}, compared to the state-of-the-art (SOTA) task-specific denoising method ADFNet \cite{ADFNet} and the SOTA general image restoration method FSNet \cite{FSNet}, ClearAIR achieves superior results, outperforming them by 0.18/0.30 dB and 0.34/0.96 dB in PSNR, respectively, at a noise level of 15 on the CBSD68 and Urban100 datasets.

\begin{figure*}[!htb]
\centering
\includegraphics[width=0.9\textwidth]{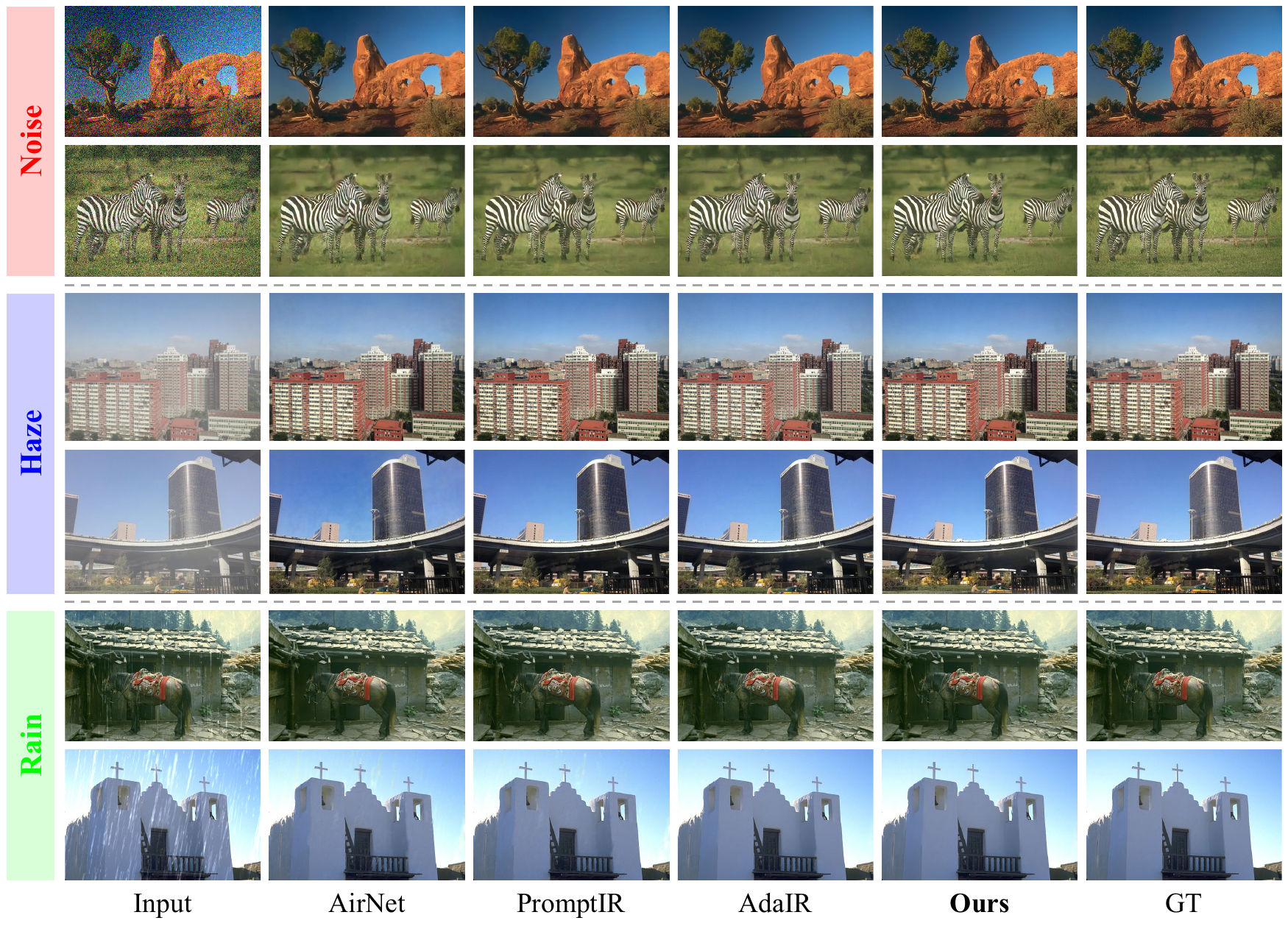}
\caption{
Visual comparisons of {ClearAIR} against state-of-the-art All-in-One methods under Single Degradation task on BSD68, SOTS-Outdoor, and Rain100L datasets (from top to bottom). Zoom-in for best view.
}
\label{fig:single_3deg}
\end{figure*}

As presented in Tab. \ref{tab:single_task_rain_haze}, ClearAIR also demonstrates the best performance in dehazing and deraining tasks. Specifically, it surpasses PromptIR by 0.60 dB and 1.48 dB in PSNR on the dehazing and deraining tasks, respectively. Even when compared to stronger single-task methods, ClearAIR still achieves improvements of 0.13 dB and 0.38 dB over DehazeFormer and DRSformer, respectively.
As illustrated in Fig. \ref{fig:single_3deg}, images restored by our method retain relatively more texture details, leading to visually more pleasing outcomes.
These results comprehensively validate the effectiveness and strong generalization ability of the proposed ClearAIR across different datasets and degradation types.

\begin{table*}[!htb] % 

  \centering
  \small

      \renewcommand\arraystretch{1.1}

    {\begin{tabular}{lccccccccc} % \resizebox{\linewidth}{!}

    \toprule[1pt]
        \toprule[0.5pt]

    \multirow{2}{*}{\bf Method} 
    & \multicolumn{3}{c}{\textbf{BSD68}}
    & \multicolumn{3}{c}{\textbf{Urban100}}  
    & \multicolumn{3}{c}{\textbf{Kodak24}}\\
    \cline{2-10} % \cmidrule(r){5-7} \cmidrule(r){8-10}

    {} 
    & 15 & 25 & 50     %& $\sigma = 15$ & $\sigma = 25$ & $\sigma = 50$
    & 15 & 25 & 50 
    & 15 & 25 & 50\\

    \midrule
    % \hline
    %************************************

    DnCNN 
    \cite{DnCNN}
    & 33.90  & 31.24  & 27.95    
    & 32.98  & 30.81  & 27.59    
    & 34.60  & 32.14  & 28.95\\

    FFDNet 
    \cite{FFDNet}
    & 33.87  & 31.21  & 27.96    
    & 33.83  & 31.40  & 28.05 
    & 34.63  & 32.13  & 28.98 \\ 

    ADFNet 
    \cite{ADFNet}
    &  34.21  &  31.60  &  28.19    
    &  34.50  &  32.13  &  28.71   
    &  34.77  & 32.22   &  29.06\\

    % \hline
    % \hline

    % MIRNet-v2 
    % \cite{MIRNet_v2}
    % & 33.66  & 30.97  & 27.66    
    % & 33.30  & 30.75  & 27.22  
    % & 34.29  & 31.81  & 28.55 \\

    DGUNet 
    \cite{DGUNet}
    & 33.85  & 31.10  & 27.92    
    & 33.67  & 31.27  & 27.94  
    & 34.56  & 32.10  & 28.91 \\ 

    Restormer 
    \cite{Restormer}  
    & 34.03  & 31.49  & 28.11    
    & 33.72  & 31.26  & 28.03   
    &  34.78  & 32.37  & 29.08\\  

    NAFNet 
    \cite{NAFNet}
    & 33.67  & 31.02  & 27.73    
    & 33.14  & 30.64  & 27.20 
    & 34.27  & 31.80  & 28.62 \\

    FSNet 
    \cite{FSNet}  
    & 34.09   &  31.55   &  28.12    
    & 33.88   &  31.31   &  28.07   
    & 34.75   &  32.38   &  29.10\\

    \midrule
    % \hline
    TAPE 
    \cite{TAPE}
    & 32.86  & 30.18  & 26.63
    & 32.19  & 29.65  & 25.87
    & 33.24  & 30.70  & 27.19\\
    
    AirNet 
    \cite{AirNet}   
    & 34.14  & 31.48  & 28.23   
    & 34.40  & 32.10  & 28.88   
    & \underline{34.81}  & \underline{32.44}  & 29.10\\

    IDR 
    \cite{IDR}
    & 34.11  & 31.60  & 28.14   
    & 33.82  & 31.29  & 28.07 
    & 34.78  & {32.42}  & \underline{29.13} \\

    PromptIR \cite{PromptIR}
    & \underline{34.34}  & \underline{31.71}  & \underline{28.49}    
    & \underline{34.77}    & \underline{32.49}  & \underline{29.39} 
    & -& -&-\\ 

    % \rowcolor{my_color}\textbf{ClearAIR}  %Perceive-IR
    % & \textbf{34.38}  & \textbf{31.74}  & \textbf{28.53}    
    % & \textbf{34.86}  & \textbf{32.55}  & \textbf{29.42} 
    % & \textbf{34.84}  & \textbf{32.50}  & \textbf{29.16}
    % \\

\bf{ClearAIR} %Perceive-IR
    & \textbf{34.39}  & \textbf{31.73}  & \textbf{28.52}    
    & \textbf{34.84}  & \textbf{32.54}  & \textbf{29.41} 
    & \textbf{34.90}  & \textbf{32.52}  & \textbf{29.27}
    \\

  \bottomrule[1pt]
  
  \end{tabular}}

  \caption{Comparison of denoising performance in the Single-task setting.} 
  \label{tab:single_denoising}
  
  % \vspace{-1em}
\end{table*}

\begin{table*}[!htb]

  \centering
  \small
  \renewcommand\arraystretch{1.1}
    {\begin{tabular}{lclc} % \resizebox{\linewidth}{!}

    \toprule[1pt]
        \toprule[0.5pt]

    \multirow{2}{*}{\bf Method} 
    & {\textbf{Dehazing}}  
    % \rowcolortwo{gray!30}
    & \multirow{2}{*}{\bf Method}
    & {\textbf{Deraining}}
        % & \cellcolor{Gray}\multicolumn{1}{c}{\textbf{Deraining}} % 这2个的空间占比是有区别的
    \\

    {}
    & {SOTS} 
    &
    & {Rain100L}
    \\
    
    \midrule
    % \hline
    DehazeNet \cite{DehazeNet}
    & 22.46/0.851 
    & UMR \cite{UMR} 
    & 32.39/0.921 
    \\

    AODNet \cite{AODNet} & 20.29/0.877 
    & MSPFN \cite{MSPFN} & 33.50/0.948
    \\

    % FDGAN \cite{FDGAN} & 23.15/0.921
    % LPNet \cite{LPNet} & 33.61/0.958
    % \\

    DehazeFormer \cite{DehazeFormer} & \underline{31.78}/\underline{0.977}  
    & DRSformer \cite{DRSformer} & \underline{38.14}/\underline{0.983}
    \\

    % general
    % \hline
    % \hline

    Restormer \cite{Restormer}  & 30.87/0.969
    & Restormer \cite{Restormer} & 36.74/0.978
    \\

    NAFNet \cite{NAFNet} & 30.98/0.970
    & NAFNet \cite{NAFNet} & 36.63/0.977
    \\

    FSNet \cite{FSNet} & 31.11/0.971
    & FSNet \cite{FSNet} & 37.27/0.980
    \\

    % All-in-One
    \midrule

    AirNet \cite{AirNet}  & 23.18/0.900
    & AirNet \cite{AirNet} & 34.90/0.977
    \\
    
    PromptIR \cite{PromptIR}  & 31.31/0.973
    & PromptIR \cite{PromptIR} & 37.04/0.979
    \\

    \bf{ClearAIR} & \bf 31.91/0.978
    & \bf{ClearAIR} & \bf {38.52/0.985}
    \\

  \bottomrule

  \end{tabular}}

  \caption{Comparison of dehazing and deraining performance in the Single-task setting.
  }
  \label{tab:single_task_rain_haze}
\end{table*}

\subsection{C. Limitation and Future Work} % y
While ClearAIR advances All-in-One Image Restoration (AiOIR) by emulating the hierarchical coarse-to-fine process of Human Visual Perception (HVP), its current design has specific limitations for real-world application. The primary weakness is the inflexibility of its fixed perception pipeline, it lacks adaptability to complex, real-world scenarios where degradation types and intensities are spatially non-uniform. This is evidenced by its performance sensitivity to the processing order in ablation studies. Furthermore, key components exhibit blind spots: the semantic guidance module struggles with inaccurate region masking under extreme low-visibility conditions, and the internal detail enhancement relies on a perceptually uninformed data augmentation strategy, which can lead to suboptimal recovery of fine details.

To address these issues,  a promising path forward is the deep integration of a Just Noticeable Difference (JND) \cite{JND} mechanism. This integration would shift the paradigm from merely mimicking the HVP workflow to embedding its fundamental perceptual laws. The envisioned JND-aware framework would employ a dynamic controller to adaptively route the restoration process based on perceptual thresholds and apply JND-weighted mechanisms to focus computational resources on perceptually critical regions. This evolution towards a perceptually-adaptive system is key to achieving significant gains in robustness and visual fidelity for complex, real-world degradations.

\begin{figure*}[!htb]
\centering
\includegraphics[width=0.9\textwidth]{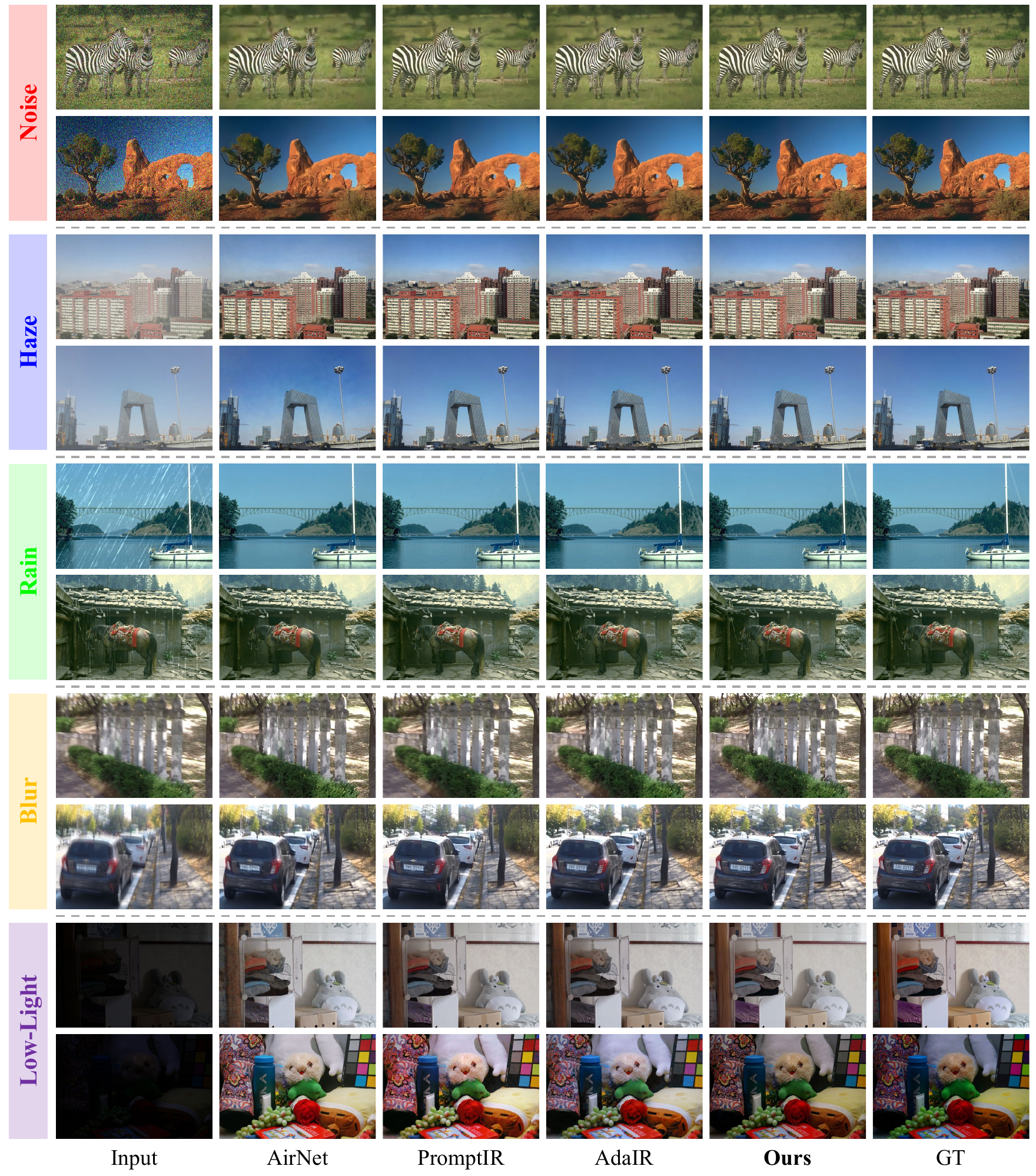}
\caption{
Visual comparisons of {ClearAIR} against state-of-the-art All-in-One methods under Five Degradation task on BSD68, SOTS-Outdoor, Rain100L, GoPro, and LOL datasets (from top to bottom). Zoom-in for best view.
}
\label{fig:five_deg}
\end{figure*}

% ----------- Supplementary Content Ends Here -----------

\end{document}